\documentclass[letterpaper]{article} 
\usepackage{aaai2026}  
\usepackage{times}  
\usepackage{helvet}  
\usepackage{courier}  
\usepackage[hyphens]{url}  
\usepackage{graphicx} 
\usepackage{natbib}  
\usepackage{caption} 
\usepackage{algorithm}
\usepackage{algorithmic}

\usepackage{amsfonts}
\usepackage{booktabs}
\usepackage{tcolorbox}
\usepackage{tabularx}
\tcbset{
  colback=gray!2,
  colframe=black!20,
  boxrule=0.3pt,
  arc=2pt,
  left=4pt,
  right=4pt,
  top=4pt,
  bottom=4pt
}
\usepackage{xcolor}
\usepackage{multirow}

\usepackage{newfloat}
\usepackage{listings}
\DeclareCaptionStyle{ruled}{labelfont=normalfont,labelsep=colon,strut=off} 
\floatstyle{ruled}
\newfloat{listing}{tb}{lst}{}
\floatname{listing}{Listing}
\title{GeoSteer: Faithful Chain-of-Thought Steering via Latent Manifold Gradients}
\author {
    Kentaro Kazama\textsuperscript{\rm 1},
    Daiki Shirafuji\textsuperscript{\rm 1},
    Tatsuhiko Saito\textsuperscript{\rm 1}
}
\affiliations {
    \textsuperscript{\rm 1}Mitsubishi Electric Corporation\\
    Kazama.Kentaro@cw.MitsubishiElectric.co.jp, Shirafuji.Daiki@ay.MitsubishiElectric.co.jp, Saito.Tatsuhiko@db.MitsubishiElectric.co.jp
}

\usepackage{bibentry}

\begin{document}

\maketitle

\begin{abstract}
Recent advances in Large Language Models (LLMs) have demonstrated remarkable progress in their reasoning capabilities, such as Chain-of-Thought (CoT). Most approaches rely on CoT rationales. Previous studies have shown that LLMs often generate logically inconsistent reasoning steps even when their final answers are correct. These inconsistencies reduce the reliability of the reasoning process. We propose GeoSteer, a manifold-based framework that improves the quality of intermediate reasoning. The method consists of: (1) constructing a CoT dataset with step-level scores, (2) training a Variational Autoencoder (VAE) model and a quality estimation model to learn a low-dimensional manifold of high-quality CoT trajectories, and (3) steering hidden states of target LLMs toward higher-quality regions in the latent space. This last step enables steering of the hidden states by following gradients along the learned manifold. It facilitates geometrically coherent steering. Evaluation experiments were conducted on the GSM8k dataset using the Qwen3 series. We evaluated performance using two metrics: answer accuracy and overall reasoning quality. GeoSteer improved the accuracy by 0.9 points and enhanced the reasoning quality by 4.5 points on average, compared with those of original LLMs. These results indicate that GeoSteer improves an effective and controllable mechanism for improving the quality of intermediate reasoning in LLMs.
\end{abstract}


\section{Introduction}

\begin{figure*}[t]
\centering
\includegraphics[width=1.99\columnwidth]{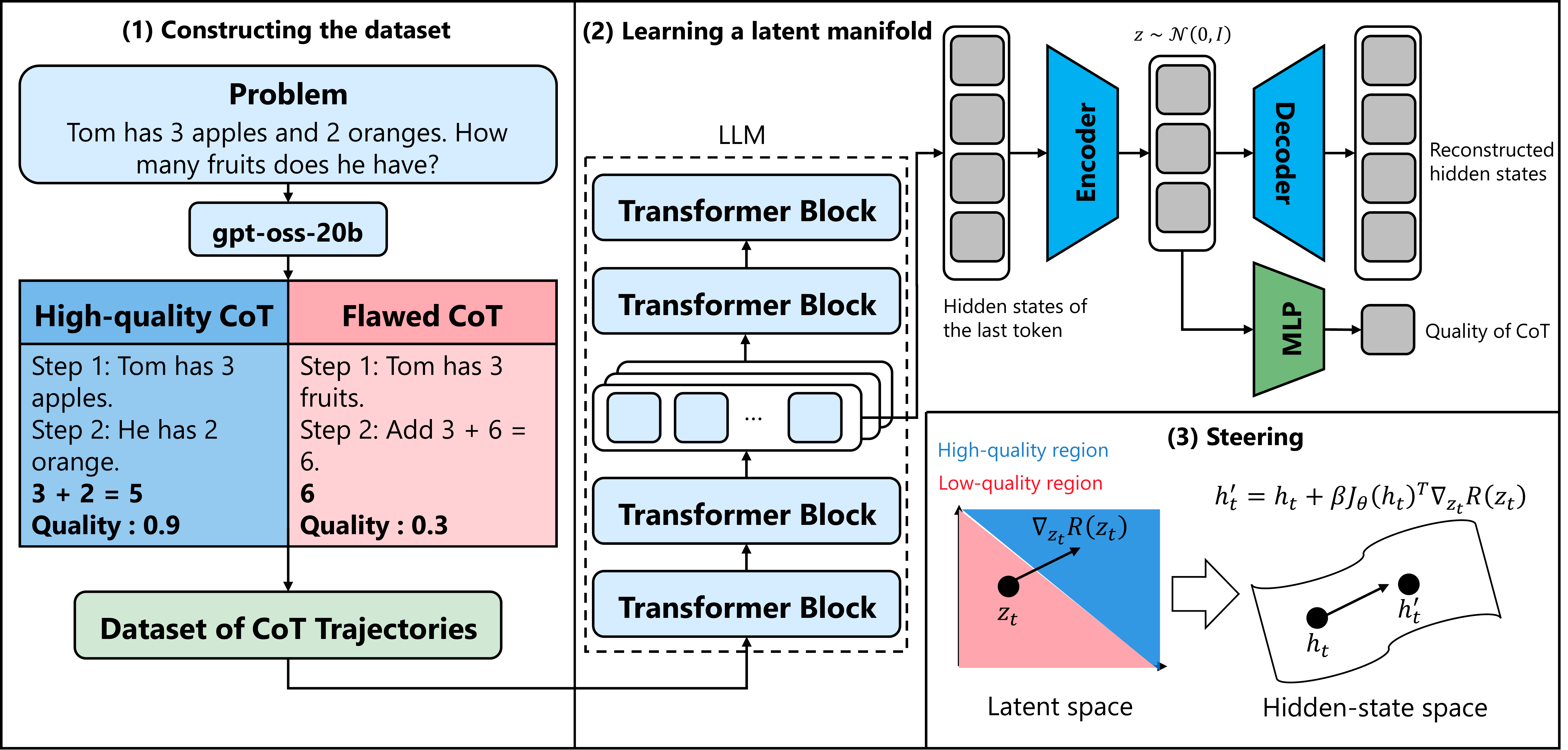}
\caption{Overview of GeoSteer. (1) A dataset of chain-of-thought (CoT) trajectories is constructed by generating multiple reasoning paths for each problem and assigning a quality score to each intermediate step. (2) A variational autoencoder (VAE) is trained on these trajectories to learn a latent manifold that captures the structural properties of high-quality reasoning. (3) During inference, the target model’s hidden states are steered by updating their latent representations toward regions associated with higher-quality reasoning.}
\label{fig:overview}
\end{figure*}

Recent progress in Large Language Models (LLMs) has led to increasingly advanced reasoning capabilities \cite{openai2023gpt4}. Techniques for generating explicit intermediate reasoning steps enable LLMs to achieve higher accuracy in downstream tasks, i.e., Chain-of-Thought (CoT) prompting \cite{wei2022cot}, Self-Consistency decoding \cite{wang2022selfconsistency}, and Tree-of-Thought \cite{yao2023treeofthoughts}. Outputting the reasoning processes facilitates the analysis of how LLMs reach their conclusions and improve the interpretability of their reasoning behavior.

Previous studies have confirmed several limitations in existing approaches, including CoT-based methods. Most approaches that train models to generate an intermediate reasoning trace (e.g., DPO \cite{rafailov2023dpo} and GRPO \cite{shao2024grpo}) require substantial preference data for an alignment and make the training process computationally expensive. To avoid these issues, activation steering methods have been proposed as alignment-free approaches that intervene directly in hidden representations. Despite addressing data and computational limitations, existing approaches still face issues. \citet{zheng2024processbench} pointed out that these errors often appear as logical inconsistencies or unnecessary sentences. \citet{yee2024unfaithfulcot} revealed that DPO-aligned models sometimes generate incorrect reasoning steps even when their final answers are correct. This inconsistency between the reasoning and the final answers reduces the reliability and interpretability of LLMs. This issue reduces the reliability of LLMs, so a reliable way to evaluate reasoning in intermediate steps is needed. As a result, errors that occur during the intermediate reasoning steps can remain unrecognized only under final-answer evaluation alone. However, most previous studies evaluate reasoning only at the step level and fail to capture the overall structural consistency of the reasoning process. These limitations indicate that existing approaches cannot provide comprehensive control over the reasoning process.

This issue of the logical consistency has recently been addressed using activation steering methods \cite{kirtania2025activation,li2025featuresteering}. \citet{kirtania2025activation} suggested an activation steering framework that increases the consistency of intermediate reasoning steps in proof generation. \citet{li2025featuresteering} proposed a training strategy that improves the overall quality of CoT reasoning without relying on high-quality CoT datasets. These results indicate that guiding the reasoning trajectory by an activation steering is an effective strategy. However, these manipulation methods operate on hidden states in a Euclidean space and rely on linear operations. This assumption may lead to degrade performance because it ignores that hidden representations are structured as points on a curved manifold. Therefore, activation steering requires operating in a manifold-based approach.

In this study, we propose a method called GeoSteer, which steers the hidden states of LLMs on a learned latent manifold. Figure~\ref{fig:overview} illustrates an overview of our proposed approach. Our method consists of three steps: (1) constructing CoT trajectories using gpt-oss-20b annotated with stepwise quality scores, which serve as high-quality reference reasoning. (2) training a VAE model on these trajectories to learn a latent manifold that captures the structure of high-quality reasoning. (3) steering the target model’s hidden states in the latent space toward regions associated with higher-quality reasoning during inference.

To capture a latent manifold, a Variational Autoencoder (VAE) is trained on high-quality reasoning trajectories.
The VAE maps hidden representations into a latent space regularized toward an approximately isotropic Gaussian distribution.
Such regularization reduces the strong anisotropy observed in the original hidden-state space, making distances and directions in the latent space more uniform and easier to manipulate.
As a result, linear updates in the latent space provide a more stable and interpretable direction for improving the quality score. Our goal is to improve the quality of CoT produced by a smaller target model by steering its hidden states during inference.

We conduct evaluation experiments on the GSM8k dataset across four LLMs in Qwen3 family. Experimental results show that our method achieves competitive accuracy in the final answers and improves the quality of the generated reasoning, i.e., its logical coherence, structural faithfulness, and stepwise consistency. These results demonstrate that GeoSteer can refine the reasoning process without decreasing the predictive performances.

Our main contributions are as follows:
\begin{enumerate}
    \item A manifold-based framework GeoSteer that adjusts hidden states in a latent space learned by the VAE to generate more logically consistent reasoning processes.
    \item A geometry-aware activation steering method that overcomes the limitations of conventional linear manipulation performed in Euclidean spaces.
    \item An empirical demonstration that GeoSteer improves the quality of CoT across multiple model parameters.
\end{enumerate}

\section{Related Works}

\subsection{Activation Steering and Representation Control}

A series of studies manipulated hidden representations to influence the behavior of large language models.  
\citet{turner2024steering} introduced a method called Activation Engineering, which applied linear activation shifts to alter factual, stylistic, or emotional attributes of model outputs.
The method identified direction vectors from contrastive examples and applied these directions to intermediate layers.

\citet{gu2024representation} extended this idea by decomposing hidden states into interpretable subspaces and associating specific components with semantic factors, such as toxicity or helpfulness.
These steering methods assumed that the activation space could be meaningfully manipulated through linear arithmetic.

Nonlinear changes in model behavior were not always expressed through such linear steering methods.  
Prior work therefore explored alternative parameterizations, including nonlinear subspace modeling \citep{meng2022locating} and low-rank adaptation of intermediate activations \citep{hu2022lora}.

Prior work also examined the distinction between inference-time and training-time control \citep{meng2022locating}.
Activation steering adjusted hidden states during inference.  
In contrast, reinforcement-learning-based alignment methods such as RLHF, DPO, and GRPO, modified model parameters through supervised or preference-based updates \citep{ouyang2022rlhf,rafailov2023dpo,shao2024grpo}.
These methods reshaped the output distribution, but did not operate on hidden representations during generation.

\subsection{Modeling and Evaluating Reasoning Trajectories}

Another line of research evaluated or improved stepwise reasoning \citep{lightman2024verify,luo2024improvemathematicalreasoninglanguage,yao2023treeofthoughts,zelikman2022star,shinn2023reflexion}.  
Chain-of-thought verification models estimated the correctness of intermediate reasoning steps \citep{lightman2024verify}.  
Process Reward Models assigned continuous quality scores to partial chains of thought and guided the generation of faithful reasoning \citep{luo2024improvemathematicalreasoninglanguage}.

Reasoning was also improved through self-refinement.  
Tree-of-Thoughts explored multiple reasoning branches and used heuristic scoring functions to select promising trajectories \citep{yao2023treeofthoughts}. 
Self-Taught Reasoner (STaR) generated additional reasoning steps and used the generated steps as additional supervision to further fine-tune the model on the same reasoning task \citep{zelikman2022star}.

Other works expanded this direction by integrating step-level evaluators with reinforcement-learning-based updates \citep{shinn2023reflexion}.
These methods operated at the text level, meaning that they evaluated or modified the sequence of generated tokens rather than the underlying hidden representations.
Many methods first generated multiple candidate reasoning trajectories and then ranked, filtered, or revised these candidates using task-specific scoring models \citep{yao2023treeofthoughts,luo2024improvemathematicalreasoninglanguage,lightman2024verify}.

Approaches that manipulated the reasoning process during generation were comparatively limited \citep{shinn2023reflexion,madaan2023selfrefine}.
Existing works mainly adjusted decoding strategies or sampling heuristics, such as verifier-driven sampling \citep{madaan2023selfrefine}.
In-process modification of hidden representations remained less explored in prior research.

\subsection{Latent Manifold Learning and Geometric Analysis}

Latent-variable models uncover low-dimensional structure in complex data.  
Variational autoencoders represent observations through latent variables with explicit priors and reconstructive decoders \citep{kingma2014vae}.  
$\beta$-VAE promotes disentangled factors in the latent representation \citep{higgins2017betavae}.  
Normalizing flows extended this framework by transforming simple latent distributions into complex densities \citep{rezende2015flows}.

Recent studies analyzed geometric patterns in neural representations of LLMs \citep{ethayarajh2019contextual,tenney2019bert}.
Hidden states have been shown to be organized into low-dimensional subspaces and to vary systematically across layers, reflecting increasing levels of semantic abstraction.
Other works reported clustering by syntactic role, semantic similarity, or task-specific difficulty \citep{liu2019linguistic,hewitt2019structural,pimentel2020information}.
These findings suggest that model representations may occupy structured regions within the activation space.

Latent manifold learning has also been explored in areas outside natural language processing, including computer vision and robotics \citep{kingma2014vae,higgins2017betavae,ha2018worldmodels}.
These approaches demonstrate that latent spaces can encode meaningful geometric structure that supports interpolation, optimization, and navigation.

\section{Preliminaries}

\subsection{Notation and Setup}

Let $x$ denote an input question, and let the model output be
\begin{equation}
    y = \left[c_0, \cdots, c_{T-1}, a\right],
\end{equation}
where $c_t$ is the $t$-th reasoning step and $a$ is the final output that answers the question.

At each step $t$, the model produces a hidden state
\begin{equation}
    h_t \in \mathbb{R}^d.
\end{equation}
The sequence $\left\{h_t\right\}_{t=0}^{T-1}$ represents the internal states generated during multi-step inference.

\subsection{Latent Representation of Hidden States}

Empirical analysis of LLMs has shown that their internal representations exhibit low-dimensional geometric structure in embedding space \cite{SaglamKassianikNelson2025,ValerianiDoimoCuturello2023,LeeWeberViegasWattenberg2025}. Motivated by these findings, we assume the existence of a latent space
\begin{equation}
    \mathcal{Z} \subset \mathbb{R}^k,\ k < d,
\end{equation}
that provides a compact coordinate system for representing hidden states. This latent space offers a reduced dimensional domain in which reasoning trajectories
\begin{equation}
    \left\{h_t\right\} \mapsto \left\{z_t\right\}
\end{equation}
can be described and compared.

\subsection{Quality of Intermediate Reasoning}

Reasoning steps in a CoT differ in how much they contribute to obtain the final answer. To express this notion, we define a scalar score
\begin{equation}
    q_t \in \left[0, 1\right],
\end{equation}
which evaluates whether the partical reasoning segment $\left\{c_0, \cdots, c_t\right\},\ (t < T)$ is sufficiently informative for deriving the correct answer.

This score reflects the coherence and correctness of reasoning up to step $t$. It provides an abstract formulation of intermediate reasoning quality independent of any specific evaluation model.

\section{Proposed Method}

We propose \textbf{GeoSteer}, a manifold-based steering framework that improves the intermediate reasoning process of LLMs. GeoSteer consists of three components: (1) constructing a dataset of high-quality and flawed CoT trajectories, (2) learning a low-dimensional latent manifold of hidden states along with a continuous quality function, and (3) steering hidden states during inference using the gradient of this quality function pulled back through the encoder.

\subsection{Construction of CoT Dataset}

Let $M_{teacher}$ be a LLM and $P$ a set of problems. For each problem $p \in P$, the teacher generates a CoT trajectory
\begin{equation}
    C_p^i = \left[c_0^i, \cdots, c_{T_i-1}^i\right],
\end{equation}
where $i \in \{\mathrm{pos}, \mathrm{neg}\}$ indicates whether the trajectory is high-quality or flawed. We generate the high-quality trajectories by prompting the LLM to produce coherent and well-structured reasoning steps. to produce coherent and well-structured reasoning steps. They typically lead to correct final outputs. Flawed trajectories contain arithmetic or logical errors while preserving natural linguistic structure.

For each prefix,
\begin{equation}
    C_{p,t}^i = \left[c_0^i, \cdots, c_{t-1}^i\right],
\end{equation}
we assign a scalar score
\begin{equation}
    q_{p,t}^i \in [0, 1].
\end{equation}
Here, partial reasoning refers to the prefix $C_{p,t}^i$, which represents the sequence of intermediate steps up to step $t$. The score evaluates whether this prefix contains sufficient information to eventually derive the correct answer. The score is produced by prompting the teacher model with the prompt shown in the Appendix.

The dataset is
\begin{equation}
    \mathcal{D} = \left\{\left(p, C_p^i, Q_p^i\right)\ |\ p \in P,\ i \in \{\mathrm{pos}, \mathrm{neg}\}\right\},
\end{equation}
where $Q_p^i = \left[q_{p,0}^i, \cdots, q_{p,T_i-1}^i\right]$ is the quality sequence aligned with the CoT steps. This dataset supervises both latent manifold learning and quality estimation.

\subsection{Learning a Latent Manifold and Quality Function}

\subsubsection{Latent Manifold of Hidden States}

Let $M_{student}$ be the model to be steered. For each trajectory $C_p^i$, we feed the CoT tokens into the student model and extract the hidden state at the final token of each prefix $C_{p,t}^i$:
\begin{equation}
    H_p^i = \left[h_{p,0}^i, \cdots, h_{p,T_i-1}^i\right],\ h_{p,t}^i \in \mathbb{R}^d,
\end{equation}
where $d$ is the dimension of the hidden states of $M_{student}$. These states represent the internal computation performed by the student model after producing step $t$.

To obtain a latent representation, we train a VAE with an encoder
\begin{equation}
    f_{\theta} : \mathbb{R}^d \rightarrow \mathbb{R}^k,
\end{equation}
and a decoder
\begin{equation}
    g_{\phi} : \mathbb{R}^k \rightarrow \mathbb{R}^d,
\end{equation}
where $k$ is the dimension of a latent space. The VAE minimizes the following objective:
\begin{equation}
    \mathcal{L}_{VAE} = \mathbb{E}_{q_{\theta}(z|h)}\left[\|g_{\phi}(z) - h\|^2\right] + \mathrm{KL}\left(q_{\theta}(z|h)\ ||\ p(z)\right),
\end{equation}
where $q_{\theta}(z|h)$ is an approximate posterior distribution, $p(z)$ is a prior distribution, and $\mathrm{KL}(\cdot)$ is a Kullback-Leibler divergence. Minimizing this objective yields a latent manifold that is both geometrically smooth and faithful to the structure of the original hidden-state space.

After training, each hidden state is mapped to a latent vector
\begin{equation}
    z_{p,t}^i = f_{\theta}\left(h_{p,t}^i\right) \in \mathcal{Z}.
\end{equation}
The latent trajectory
\begin{equation}
    Z_p^i = \left[z_{p,0}^i, \cdots, z_{p,T_i-1}^i\right]
\end{equation}
forms a low-dimensional geometric structure that reflects the progression of reasoning in the student model.

\subsubsection{Latent Quality Function}

We learn a differentiable scoring function
\begin{equation}
    R_{\psi} : \mathcal{Z} \rightarrow \mathbb{R},
\end{equation}
which predicts the reasoning-quality score from latent vectors. Given training pairs $\left(z_{p,t}^i, q_{p,t}^i\right)$, we fit the function by minimizing
\begin{equation}
    \mathcal{L}_{score} = \frac{1}{|\mathcal{D}|}\sum_{p,i,t}\left(R_{\psi}(z_{p,t}^i) - q_{p,t}^i\right)^2.
\end{equation}
The gradient
\begin{equation}
    \nabla_{z}R_{\psi}(z)
\end{equation}
indicates the direction in latent space that improves the quality of intermediate reasoning.

\subsection{Steering of Hidden States}

To apply latent gradients to hidden states, we compute the pullback through the encoder. Let $J_{\theta}(h_t)$ denote the Jacobian of $f_{\theta}$ at $h_t$. By the chain rule, we obtain the following:
\begin{equation}
    \nabla_{h_t}R_{\psi}(z_t) = J_{\theta}(h_t)^\top\nabla_{z_t}R_{\psi}(z_t).
\end{equation}

During inference, the student model generates hidden states $h_t$ autoregressively. For each step, we first compute the latent representation
\begin{equation}
    z_t = f_{\theta}(h_t),
\end{equation}
an obtain the pullback gradient
\begin{equation}
    \nabla_{h_t}R_{\psi}(z_t).
\end{equation}
The hidden state is updated using a normalized gradient
\begin{equation}
    h_t' = h_t + \beta\frac{\nabla_{h_t}R_{\psi}(z_t)}{\left\|\nabla_{h_t}R_{\psi}(z_t)\right\|},
\end{equation}
where $\beta$ is a hyperparameter that controls the intensity of steering at each reasoning step. The updated state $h_t'$ is used to generate the next token. Repeating this over all steps produces a reasoning trajectory guided toward higher-quality regions of the latent manifold while preserving the student model's linguistic behavior.

\section{Experiments}

In this section, we evaluate the effectiveness of the proposed method through both quantitative and qualitative experiments. We first describe the experimental setup, then present results and provide a case study illustrating latent-space trajectories. The goal is to examine how latent-space steering influences both the reasoning quality and the final-answer accuracy of LLMs.

\subsection{Experimental Setup}
\label{sec:experimental_setup}

\subsubsection{Models}

We use Qwen models: Qwen3-0.6B, Qwen3-1.7B, Qwen3-4B, and Qwen3-8B \cite{qwen2025qwen3}. Each model generates CoT using an identical prompt template. We evaluate two settings: a baseline without steering, and a steered version that modifies hidden states through latent-space updates. All steering operations use the latent manifold learned by the VAE.

As a teacher model, we employ gpt-oss-20b \cite{openai2025gptoss120bgptoss20bmodel}, which produces high-quality CoT trajectories. These trajectories are used only during training of the VAE and the regressor $R_{\psi}$.

\subsubsection{CoT Dataset}

We evaluate our methods with the GSM8k dataset \cite{cobbe2021gsm8k}, a benchmark for arithmetic and logical reasoning. Each data consists of a natural language question paired with a numerical answer derivable through step-by-step reasoning. The teacher model generates two types of CoT for each training example:
\begin{enumerate}
    \item High-quality CoT, which follows correct and coherent reasoning steps.
    \item Low-quality CoT, which contains incorrect, incomplete, or inconsistent reasoning.
\end{enumerate}
Both types are converted into hidden states $\{h_t\}$, and then into latent representations $\{\mu_t\}$. These pairs $(\mu_t, q_t)$ constitute the training dataset for the VAE and the regressor, where $q_t$ denotes the quality score assigned by the teacher model.

\subsubsection{Evaluation Dataset}

We evaluate our method in the GSM8k test set in the 4-shot setting. The four samples are randomly chosen from the GSM8k test set. The few-shot prompt is shown in Appendix.

\subsubsection{Evaluation Metrics}

To independently assess the accuracy of the final answer and reasoning quality, we employ two complementary metrics.
\begin{enumerate}
    \item Exact Match (EM): EM measures whether the final predicted answer matches the ground truth exactly.
We report the difference between the baselines and the steered models.
    \item Pairwise Win Rate : For each test example, GPT-4o compares the CoT generated by the baseline and steered models. The judge selects the CoT that is superior in clarity, coherence, and reasoning quality. The win rate is the percentage of cases in which the steered model is preferred. The prompt for pairwise win rate is provided in the Appendix.
\end{enumerate}

\subsection{Experimental Results}
\label{sec:experimental_results}

\subsubsection{Exact Match Accuracy}

\begin{table*}[t]
\centering
\footnotesize
\begin{tabular}{r|rr|rr|rr|rr}
\toprule
\multirow{2}{*}{$\beta$}
& \multicolumn{2}{c|}{\textbf{Qwen3-0.6B}} 
& \multicolumn{2}{c|}{\textbf{Qwen3-1.7B}}
& \multicolumn{2}{c|}{\textbf{Qwen3-4B}}
& \multicolumn{2}{c}{\textbf{Qwen3-8B}} \\
\cmidrule(lr){2-9}
& Baseline & Steered & Baseline & Steered & Baseline & Steered & Baseline & Steered \\
\midrule
1   & 60.0 & 58.7 & 82.3 & \textbf{82.4} & 90.6 & 90.5 & 90.7 & 90.4 \\
10  & 60.0 & 60.0 & 82.3 & \textbf{82.9} & 90.6 & 90.5 & 90.7 & 90.6 \\
50  & 60.0 & 58.5 & 82.3 & \textbf{83.1} & 90.6 & 90.3 & 90.7 & 90.4 \\
100 & 60.0* & 55.0 & 82.3 & \textbf{83.5} & 90.6 & 89.5 & 90.7 & \textbf{90.8} \\
125 & 60.0* & 52.0 & 82.3 & \textbf{83.5} & 90.6 & 89.8 & 90.7 & \textbf{91.3} \\
150 & 60.0* & 50.9 & 82.3 & \textbf{84.9}* & 90.6 & 89.8 & 90.2 & \textbf{91.3} \\
200 & 60.0* & 46.2 & 82.3 & \textbf{84.1}* & 90.6 & 89.9 & 90.7 & \textbf{91.4} \\
300 & 60.0* & 28.7 & 82.3 & \textbf{84.7}* & 90.6* & 88.9 & 90.7 & \textbf{91.3} \\
\bottomrule
\end{tabular}
\caption{
Exact Match (\%) for different steering strengths $\beta$ across Qwen3 model scales. Bold numbers indicate improvements over the baseline. * shows statistical significance at $p < 0.05$ on the McNemar test.
}
\label{tab:em_beta_sweep}
\end{table*}

\begin{table*}[t]
\centering
\footnotesize
\begin{tabular}{r|rr|rr|rr|rr}
\toprule
\multirow{2}{*}{$\beta$}
& \multicolumn{2}{c|}{\textbf{Qwen3-0.6B}} 
& \multicolumn{2}{c|}{\textbf{Qwen3-1.7B}}
& \multicolumn{2}{c|}{\textbf{Qwen3-4B}}
& \multicolumn{2}{c}{\textbf{Qwen3-8B}} \\
\cmidrule(lr){2-9}
& Baseline & Steered & Baseline & Steered & Baseline & Steered & Baseline & Steered \\
\midrule
1   & 8.42 & \textbf{11.52} & 1.82* & \textbf{3.71}* & 0.45 & \textbf{1.90}* & 0.38 & \textbf{1.90}* \\
10  & 10.01 & \textbf{15.31} & 1.90* & \textbf{4.78}* & 0.68 & \textbf{2.81}* & 0.53 & \textbf{2.05}* \\
50  & 15.92 & \textbf{19.86} & 2.81* & \textbf{6.82}* & 1.97 & \textbf{3.11} & 1.06 & \textbf{3.03}* \\
100 & 21.53 & \textbf{23.20} & 3.64 & \textbf{7.51}* & 3.11 & 3.11 & 1.21 & \textbf{2.81}* \\
125 & \textbf{24.18}* & 19.41 & 4.40 & \textbf{7.66}* & 3.26 & 2.81 & 1.29 & \textbf{4.25}* \\
150 & \textbf{26.84}* & 20.92 & 4.32 & \textbf{7.66}* & 3.03 & \textbf{3.64} & 1.52 & \textbf{8.04}* \\
200 & \textbf{33.21}* & 20.24 & 5.53 & \textbf{6.97} & 2.81 & \textbf{3.64} & 0.99 & \textbf{4.70}* \\
300 & \textbf{59.89}* & 15.62 & 5.00 & \textbf{8.49}* & 4.93 & 3.41 & 1.82 & \textbf{5.16}* \\
\bottomrule
\end{tabular}
\caption{
Pairwise win rate (\%) for different steering strengths $\beta$ across Qwen3 model scales.  
Values show the percentage of comparisons where each model's output was preferred by GPT-4o. Bold numbers indicate improvements over the baseline. * shows statistical significance at $p < 0.05$ on the McNemar test.
}
\label{tab:pairwise_beta_sweep}
\end{table*}

Table ~\ref{tab:em_beta_sweep} reports Exact Match (EM) for all model sizes and steering strengths.
The steered models showed small changes in EM at different values of $\beta$.

For Qwen3-1.7B, EM improved from 82.3 to 84.9 at $\beta=150$, which was the largest gain observed.
Qwen3-8B also showed moderate improvements when $\beta$ was between 125 and 150.
Qwen3-4B remained stable at all $\beta$ values.
In contrast, Qwen3-0.6B was sensitive to the steering strength, and EM dropped sharply when $\beta > 100$.

These results showed that latent-space steering preserved factual accuracy at most operating points, and occasionally improved it for medium to large models.

\subsubsection{Pairwise Evaluation}

Table ~\ref{tab:pairwise_beta_sweep} shows the pairwise win rate obtained using GPT-4o as a judge.
The steered models were consistently preferred over the baselines.

For Qwen3-0.6B, the win rate increased from 10.01\% to 15.31\% at $\beta=10$.
For Qwen3-1.7B, the steered model showed a consistent improvements in win rate at every value of tested $\beta$.
Qwen3-4B and Qwen3-8B also achieved higher win rates, although the gains were smaller than those of smaller models.

These results indicated that steering improved the perceived quality of CoT reasoning, even when EM remained unchanged.

\subsubsection{Effect of Steering Strength}

We further investigated how different steering strengths influence EM and pairwise performance.
The trends varied between model sizes.

Qwen3-1.7B and Qwen3-8B achieved their best EM at $\beta\in\{125,150\}$.
Qwen3-0.6B showed improvements in pairwise evaluation at moderate $\beta$ but degraded sharply at high $\beta$.
Across all models, pairwise win rates increased for a wide range of $\beta$, with the strongest gains observed for Qwen3-0.6B at $\beta\in\{100,125,150,200\}$. These observations suggested that the optimal $\beta$ depends on the model capacity.

\section{Discussion}

\subsection{Behavior of Latent-Space Steering}

\begin{table*}[t]
    \centering
    \small
    \begin{tcolorbox}
        \begin{tabularx}{\textwidth}{@{}lX@{}}
            \textbf{Problem:}             &
            An 8-year old child wants to buy a toy car which costs \$12. He already has \$4 savings. How many days will it take him to save the remaining amount of money if he promises to save \$2 daily from his allowance?                                                                                                                                    \\\\[6pt]

            \textbf{Answer A (Baseline):} &
            \verb|<think>|

Okay, let's see. So the problem is about Julie saving money for a toy car. The car costs \$12, and she already has \$4.

So the remaining money she needs is \$12 minus \$4, which is \$8. She wants to save \$2 every day.

To figure out how many days it will {\color{red} take}, we divide the remaining amount by the daily savings.

So \$8 divided by \$2 is 4 days. That means it will take her 4 days to save the remaining \$8. The final answer should be 4.

\verb|</think>|

The final answer is 4. \\\\[6pt]

            \textbf{Answer B (Steered):}  &
            \verb|<think>|

Okay, let's see. So, Julie wants to buy a toy car that costs \$12. She already has \$4 saved. That leaves her with \$12 - \$4 = \$8 to save. She plans to save \$2 each day.

To find out how many days it will {\color{red} take}, we divide the remaining amount by the daily savings.

So, \$8 divided by \$2 per day equals 4 days. Therefore, it will take her 4 days to save the remaining money.

            \verb|</think>|

            The final answer is 4.
        \end{tabularx}
    \end{tcolorbox}
    \caption{
        Sample problem and its answers correspond to Figure ~\ref{fig:visualization_of_trajectory}. The red words indicate the token numbered 64.
    }
    \label{tab:case_cot_for_visualization}
\end{table*}

\begin{figure}[t]
\centering
\includegraphics[width=1.2\columnwidth,clip]{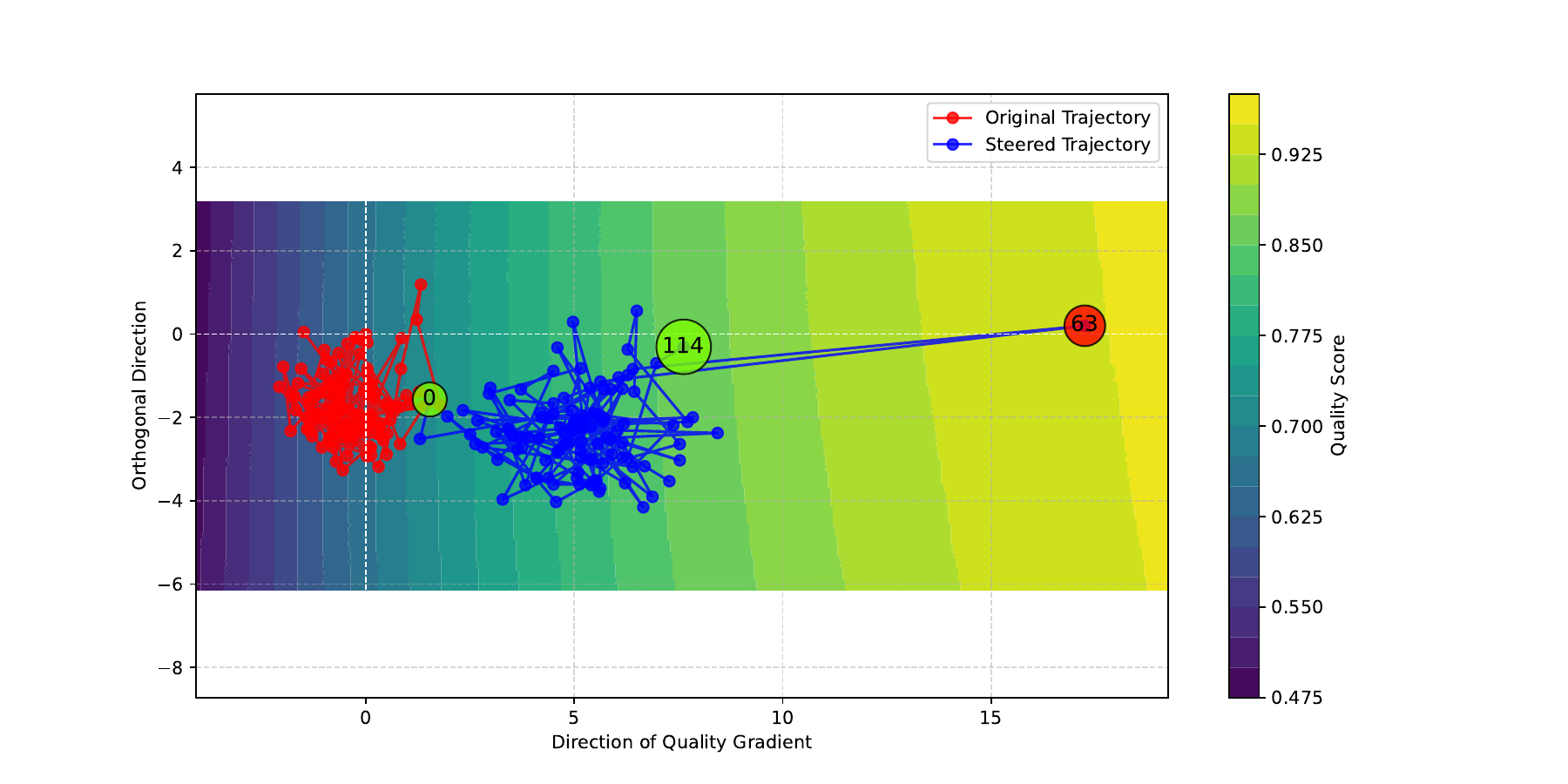}
\caption{Latent reasoning trajectories projected onto a 2D plane defined by the quality-gradient direction (x-axis) and its orthogonal direction (y-axis). Red line indicates the baseline trajectory, and blue line indicates the steered trajectory.}
\label{fig:visualization_of_trajectory}
\end{figure}

To examine how latent-space steering modifies the internal reasoning process,
we analyzed latent trajectories projected onto the two-dimensional space.
Figure~\ref{fig:visualization_of_trajectory} shows the latent states $\mu_t$ produced for the problem in Table~\ref{tab:case_cot_for_visualization}, projected onto a two-dimensional plane defined by the quality-gradient direction (x-axis) and its orthogonal axis (y-axis).

The baseline trajectory stayed within a region whose average quality score was around 0.65. It showed minor directional shifts throughout the reasoning process.
In contrast, the steered trajectory showed a sharp transition around token 63, which corresponded to the word ``take'' in Table~\ref{tab:case_cot_for_visualization}. At this point, the average quality score increased from approximately 0.65 to around 0.78.
This point marked a semantic shift from the problem description to the computation phase, where the predicted quality gradient became steep in the latent space.

An important observation was that this sudden movement occurred despite the fact that both generations produced the same surface text.
The steering effect did not arise from differences in output tokens. It reflected a change in the internal representation induced by the latent quality gradient.
The trajectory was pulled toward a higher-quality region precisely at the moment when the model began to perform arithmetic reasoning.

After the transition, the steered trajectory remained near the high-quality region and progressed smoothly toward the final state. In this example, the high-quality region corresponded to quality scores around 0.75–0.80.
Consequently, the steered CoT became slightly more concise while still preserving all essential reasoning steps, whereas the baseline CoT included a few additional explanatory phrases even though both models reached the same final answer (“4 days”).
The latent-space trajectory thus revealed a more coherent and stable reasoning pattern for the steered model.

This example demonstrated that the proposed method was capable of controlling the reasoning trajectory, rather than altering the generated text.
In particular, the most significant influence of latent-space steering was observed at the semantically important boundaries, such as the transition into the computation phase. It guided the outputs of the model to the regions corresponding to higher-quality CoTs.
These observations suggest that our method improves the overall structural consistency of the reasoning process, beyond the surface-level similarity of the generated text.

\subsection{Relationship Between CoT Quality and Final Accuracy}

\begin{figure}[t]
\centering
\includegraphics[width=1.0\columnwidth,clip]{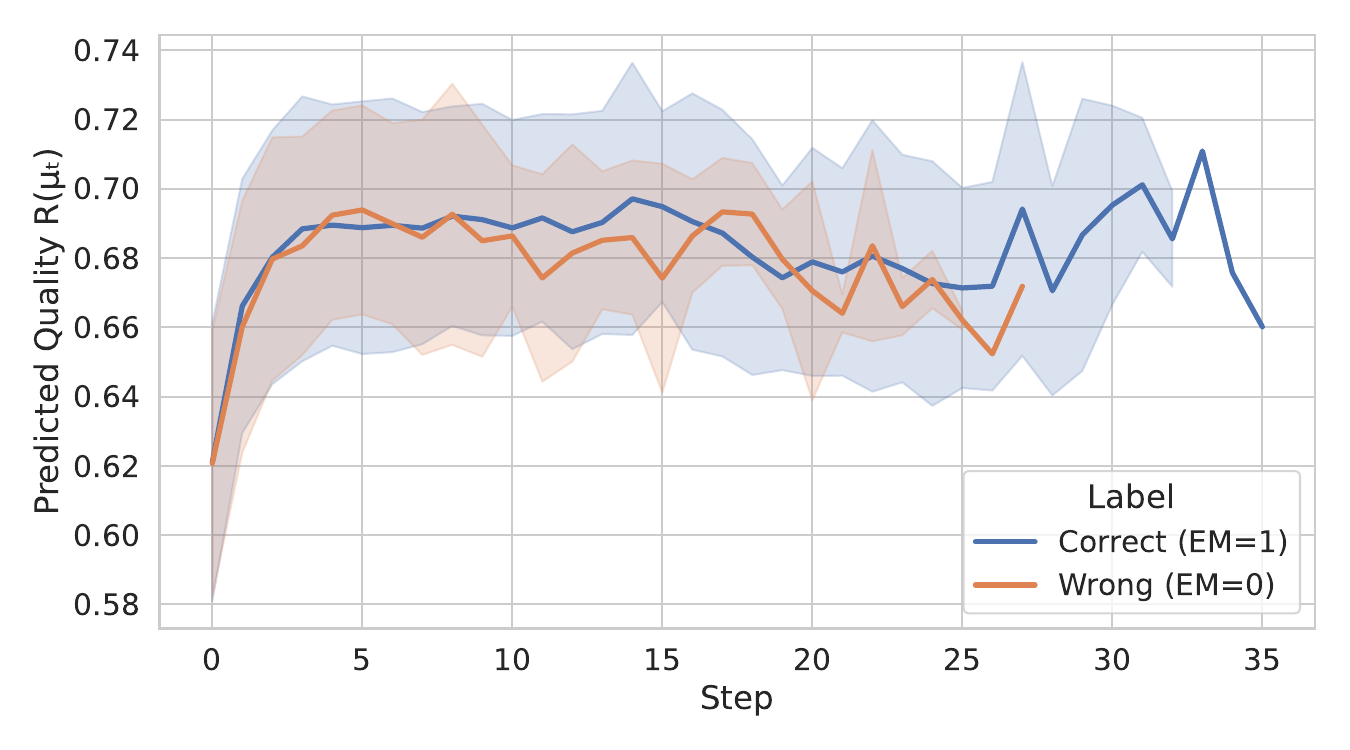}
\caption{Step-wise reasoning-quality scores averaged over correct (EM=1) and incorrect (EM=0) solutions.}
\label{fig:visualization_quality_em}
\end{figure}

Figure~\ref{fig:visualization_quality_em} plots the step-wise reasoning-quality curves averaged over correct (EM=1) and incorrect (EM=0) solutions.
The two curves exhibited similar shapes: the quality increased in the early steps and then stabilized, and their variance bands substantially overlapped throughout the trajectory.
This indicates that the latent-space quality estimator primarily captures local coherence of intermediate reasoning rather than global correctness.

We also observed that incorrect examples tended to contain fewer reasoning steps.
This resulted in a smaller variance toward the end of the trajectory.
This pattern suggests that the model often terminates early once it enters an unproductive reasoning path, even though the step-by-step logic of the partial reasoning remains superficially coherent.
The similarity of the two curves, therefore does not imply similar final outcomes, but rather reflects similarities in how partial reasoning is structured.

Furthermore, we found cases in which a trajectory maintained high quality scores across most intermediate steps but still produced an incorrect final answer due to a small arithmetic mistake at the end.
In contrast, some correct answers arose from trajectories that included low-quality regions or fluctuations in the middle of the reasoning.
These observations indicate that CoT quality and EM capture different aspects of model behavior:
EM depends heavily on the correctness of the final computation, whereas the latent-space score evaluates the structural coherence of partial reasoning.

In general, Figure~\ref{fig:visualization_quality_em} shows that intermediate reasoning quality is useful, but it is not sufficient to predicting the final answer.
This highlights the need for evaluation metrics that capture global logical consistency rather than rely solely on local coherence.

\subsection{Future Works}

This work provided a step toward controlling the internal reasoning process of LLMs through latent-space geometry.
Some directions remain for future exploration.

The development of more reliable metrics to evaluate the quality of CoT reasoning processes is essential.
The step-wise quality score used in this study captures local coherence but does not fully reflect global logical consistency.
To evaluate such global properties, more comprehensive metrics are needed. These assess multi-step causal correctness or trajectory-level validity.

A deeper analysis of latent trajectories across diverse model families is needed.
Our experiments focused primarily on the Qwen series, which explicitly supports CoT generation.
It is necessary to assess if similar trends seen in Qwen emerge in other LLM architectures, such as LLaMA, Mistral, and DeepSeek. This evaluation will verify whether the proposed method is generally applicable to LLMs.

\section{Conclusion}

We proposed the method GeoSteer, a latent-space framework for adjusting the internal reasoning process of large language models.
Our method computed quality gradients in a VAE latent space and pulled them back into the hidden-state space to provide geometry-aware updates.
The steering operation guided the reasoning trajectories toward regions associated with high-quality CoTs.

Our experiments demonstrated that latent-space steering improves the coherence of intermediate reasoning without harming final-answer accuracy.
The analysis showed that our steering method altered the internal structure of reasoning, often at key transition points. Trajectory analyses confirmed this change, even when the generated text was nearly identical to the baseline.
The score curves revealed that intermediate reasoning quality and final-answer correctness are related but not equivalent.
These findings indicate the importance of the evaluation on the consistency of reasoning itself rather than solely on the final answers.

In the future, we plan to construct evaluation metrics because the quality score captures only local coherence and does not fully reflect global logical consistency.
We also plan to experiment with models from various series to analyze differences in reasoning trajectories.

\section{Liminations}

Our method relies on the CoT dataset generated by gpt-oss-20b, which effectively sets the upper bound of the achievable quality of reasoning.
The latent manifold and the quality function are therefore bounded by the coverage of gpt-oss-20b and may not fully reflect the structure of higher-quality reasoning.  
This dependency limits the achievable improvement on downstream tasks.

The geometric assumptions underlying GeoSteer remain partially heuristic.  
The VAE imposes a continuous latent space that enables smooth steering, but it does not guarantee that the coordinates correspond to a well-defined reasoning manifold.  
The relationship between gradients in latent space and the true geometry of hidden-state trajectories is not yet theoretically established.  
A more rigorous analysis is required to clarify the geometric validity of the method and to determine whether the learned manifold faithfully reflects the structure of high-quality reasoning.




\bibliography{aaai2026}


\clearpage

\section{Appendix}
\label{appendix:prompts}

\subsection{Prompt for CoT Quality Scoring (gpt-oss-20b)}
\label{appendix:partial-cot}
This appendix provides the prompt for scoring the quality of CoT.

\begin{tcolorbox}[title=CoT Quality Scoring Prompt,colback=white,colframe=black!50]
{\footnotesize
You are an expert reasoning evaluator.

You will rate how logically consistent and correct the reasoning is
with respect to solving the following problem.

Problem:

\verb|{problem_text}|

Reasoning prefix:

\verb|{prefix_text}|

Give a single numeric score between 0.0 and 1.0 where:

- 1.0 = perfectly correct reasoning leading to a valid conclusion,

- 0.0 = completely wrong or illogical reasoning,

- values in between indicate partial correctness.

Respond only with the numeric score.
}
\end{tcolorbox}

\subsection{Prompt for Pairwise Win-Rate Evaluation (GPT-4o)}
\label{appendix:pairwise}
This appendix introduces the prompt for evaluating Pairwise Win-Rate on experiments.

\begin{tcolorbox}[title=Pairwise Win-Rate Prompt,colback=white,colframe=black!50]
{\footnotesize
You are an expert evaluator of reasoning quality and answer correctness. 

You will be given two responses, Response A and Response B, to the same problem. 

Your job is to judge which one is better overall in terms of:

  1. Correct final answer (most important)

  2. Logical coherence and clarity of reasoning steps

  3. Conciseness (less irrelevant detail)

  4. Self-consistency (no arithmetic errors, no contradictions)

**Instructions:**

1. Extract the final answer from each response and compare it to the ground truth. 
   If one answer is correct and the other is incorrect, the correct one *wins*.

2. If both answers are correct, compare reasoning: choose the one that is more logically coherent, clearer, and more concise.

3. If both are incorrect, compare how close each reasoning is to a correct reasoning path (e.g., minor arithmetic slip vs completely wrong approach).

4. If both responses are equivalent in correctness and reasoning quality (no clear winner), score as a **Tie**.

5. Provide exactly one of the following tokens as a following format: \verb|`<<A>>`|, \verb|`<<B>>`|, or \verb|`<<Tie>>`|.

6. After that, provide a short justification (1-3 sentences) for your decision.

**Inputs:**

Problem: \verb|{problem}|

Response A: \verb|{answer_a}|

Response B: \verb|{answer_b}|

Ground truth answer: \verb|{ground_truth}|

Judge: 
}
\end{tcolorbox}

\subsection{Few-shot Prompt for Steering and Evaluation}
This appendix presents the few-shot prompt to input the target model for steering and evaluation.

\label{appendix:fewshot}
\begin{tcolorbox}[title=Few-shot Prompt,colback=white,colframe=black!50]
{\footnotesize
\verb/<|im_start|>/system

You are a concise math assistant. Use \verb|<think>|\verb|</think>| for internal reasoning,

but after \verb|</think>| only output the final numeric answer in the format:

'The final answer is \verb|<number>|.'

\verb/<|im_end|>/

\verb/<|im_start|>/user

Natalia sold clips to 48 of her friends in April, and then she sold half as many clips in May. How many clips did Natalia sell altogether in April and May?

\verb/<|im_end|>/

\verb/<|im_start|>/assistant

The final answer is 72.

\verb/<|im_end|>/

\verb/<|im_start|>/user

Weng earns \verb|$|12 an hour for babysitting. Yesterday, she just did 50 minutes of babysitting. How much did she earn?

\verb/<|im_end|>/

\verb/<|im_start|>/assistant

The final answer is 10.

\verb/<|im_end|>/

\verb/<|im_start|>/user

Betty is saving money for a new wallet which costs \verb|$|100. Betty has only half of the money she needs. Her parents decided to give her \verb|$|15 for that purpose, and her grandparents twice as much as her parents. How much more money does Betty need to buy the wallet?

\verb/<|im_end|>/

\verb/<|im_start|>/assistant

The final answer is 5.

\verb/<|im_end|>/

\verb/<|im_start|>/user

Julie is reading a 120-page book. Yesterday, she was able to read 12 pages and today, she read twice as many pages as yesterday. If she wants to read half of the remaining pages tomorrow, how many pages should she read?

\verb/<|im_end|>/

\verb/<|im_start|>/assistant

The final answer is 42.

\verb/<|im_end|>/
}
\end{tcolorbox}

\end{document}